\pdfoutput=1
\documentclass[letterpaper]{article}
\usepackage[totalwidth=480pt, totalheight=680pt]{geometry}
\usepackage{authblk}
\usepackage{amsmath}

\usepackage[T1]{fontenc}
\usepackage{amsfonts}
\usepackage{graphicx}
\usepackage{subcaption} 
\usepackage{epstopdf}
\usepackage{enumitem}
\usepackage{algorithmic}
\ifpdf
  \DeclareGraphicsExtensions{.eps,.pdf,.png,.jpg}
\else
  \DeclareGraphicsExtensions{.eps}
\fi

\usepackage{amsthm}
\usepackage{thmtools, thm-restate}

\theoremstyle{plain}
\newtheorem{theorem}{Theorem}
\newtheorem{lemma}[theorem]{Lemma}

\newtheorem{assumption}[theorem]{Assumption}

\theoremstyle{definition}
\newtheorem{definition}[theorem]{Definition}
\newtheorem{example}[theorem]{Example}

\theoremstyle{remark}

\usepackage{amssymb}
\usepackage{amsopn}
\usepackage{xcolor}
\DeclareMathOperator{\diag}{diag}
\usepackage{macro}

\usepackage{hyperref}
\usepackage{cleveref}

\begin{document}

\newcommand{\independent}{\mathrel{\perp\!\!\!\perp}}
\newcommand{\notindependent}{\mathrel{\not\!\perp\!\!\!\perp}}
\newcommand\relatedversion{}
\renewcommand\relatedversion{\thanks{The full version of the paper can be accessed at \protect\url{https://arxiv.org/abs/0000.00000}}} 

\title{\Large 
Causal Structure and Representation Learning \\with Biomedical Applications
\footnote{Forthcoming in the Proceedings of the \emph{International Congress of Mathematicians 2026, EMS Press}. Both authors contributed
equally to this work.}
}
\author[1,*]{Caroline Uhler}
\author[1,*]{Jiaqi Zhang}
\affil[1]{Department of Electrical Engineering and Computer Science, Massachusetts Institute of Technology}
\affil[*]{Emails: culer@mit.edu, viczhang@mit.edu}

\date{}

\maketitle




\begin{abstract}  
Massive data collection holds the promise of a better understanding of complex phenomena and, ultimately, better decisions. Representation learning has become a key driver of deep learning applications, as it allows learning latent spaces that capture important properties of the data without requiring any supervised annotations. Although representation learning has been hugely successful in predictive tasks, it can fail miserably in causal tasks including predicting the effect of a perturbation/intervention. This calls for a marriage between representation learning and causal inference. An exciting opportunity in this regard stems from the growing availability of multi-modal data (observational and perturbational, imaging-based and sequencing-based, at the single-cell level, tissue-level, and organism-level). We outline a statistical and computational framework for causal structure and representation learning motivated by fundamental biomedical questions: how to effectively use observational and perturbational data to perform causal discovery on observed causal variables; how to use multi-modal views of the system to learn causal variables; and how to design optimal perturbations.   
\end{abstract}

\section{Introduction.}\label{sec:intro}
Causality is concerned with understanding the underlying mechanisms that govern a system. It often centers on fundamental questions such as: What is the underlying data-generating process that can explain observed phenomena? What are the cause-effect relationships among the observed variables? How do the observed variables change under specific interventions/perturbations? And what are optimal interventions/perturbations in order to move the system to a desired state? Addressing such questions requires moving beyond correlations and understanding causal mechanisms.

\vspace{0.2cm}

\begin{figure}[ht]
    \centering
    \begin{subfigure}{0.4\linewidth}
        \centering
        \includegraphics[width=.7\linewidth]{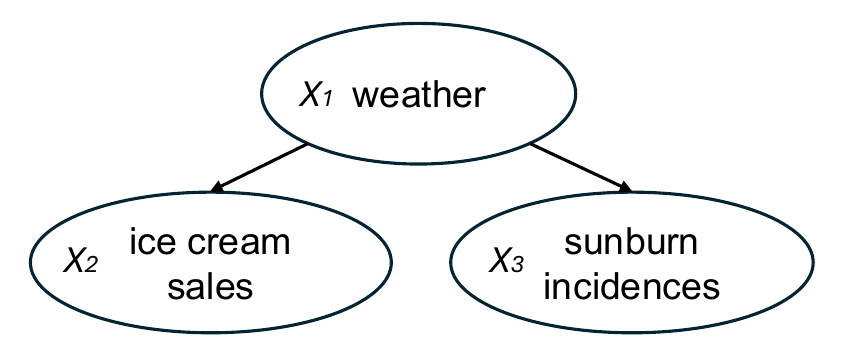}
        \caption{Ice cream sales and sunburn incidences.}\label{fig:1a}
    \end{subfigure}
    \begin{subfigure}{0.5\linewidth}
        \centering
        \includegraphics[width=.595\linewidth]{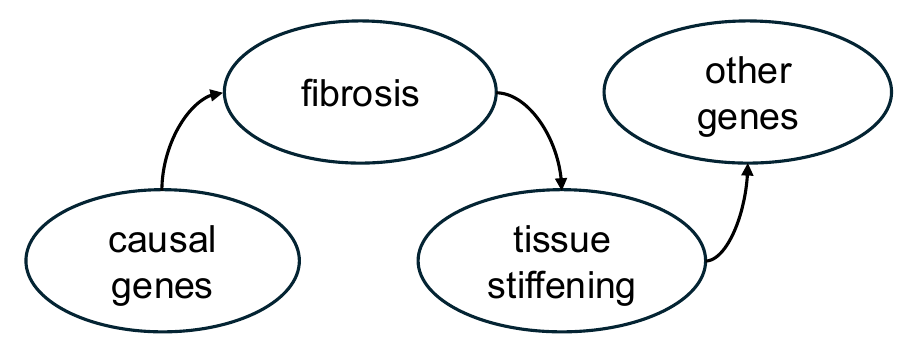}
        \caption{Fibrosis and causal genes/proteins.}\label{fig:1b}
    \end{subfigure}
    \caption{Illustrative examples and their respective causal graphs.}\label{fig:1}
\end{figure}

\noindent\textbf{Examples:} For illustration, consider the example of ice cream sales and sunburn incidences (\cref{fig:1a}). Although there is a strong positive correlation between the two, one does not cause the other. Instead, a third variable, sunny weather, is a common cause of both variables. This simple example illustrates the importance of understanding causality, as misinterpretation could lead to ineffective interventions, such as banning ice cream to prevent sunburns.
Another example is fibrosis, which is responsible for up to 45\% of deaths in the industrialized world~\cite{younesi2024fibroblast}. Fibrosis is associated with changes in many genes/proteins. However, only a subset of genes/proteins are causal factors of fibrosis, while various other genes/proteins may be affected by tissue stiffening and thus downstream of fibrosis (\cref{fig:1b}).
Effective therapies require disentangling \emph{upstream causal} genes/proteins, which represent potential \emph{therapeutic targets}, from \emph{downstream biomarkers} of the disease.

\vspace{0.2cm}

\noindent\textbf{Causal DAG:} A causal system is commonly represented by a directed acyclic graph (DAG), where each node is associated with a random variable and each directed edge represents a direct causal relationship~\cite{lauritzen1996graphical,spirtes2000causation,pearl2009causality}. While extensions to cyclic models have been developed~\cite{richardson2013discovery,mooij2011causal,lacerda2012discovering,peters2017elements}, acyclicity has traditionally been assumed since causality acts forward in time, and we here concentrate on DAGs. Let $\cG=([p],E)$ be a DAG with nodes $[p]:=\{1,\dots,p\}$ and directed edges $E$. Each node $i$ in $\cG$ is associated with a random variable $X_i$ and an edge $i\rightarrow j$ in $\cG$ indicates that $X_i$ is a direct cause of $X_j$. 
The \emph{Markov property} relates the joint distribution of $\bX=(X_1,...,X_p)^\top$ to $\cG$, defined as follows.

\begin{definition}\label{def:markov}
    A joint distribution $\bbP$ is \emph{Markov} with respect to a DAG $\cG$ if it factorizes according to
\begin{equation}\label{eq:markov-factorization}
\bbP(X_1,\dots,X_p) = \prod_{i=1}^p \bbP(X_i\mid X_{\Pa_\cG(i)}),
\end{equation}
where $\Pa_\cG(i):=\{j\in [p] : i\leftarrow j \in E\}$ denotes the parents of $i$ in $\cG$. 
\end{definition}
This factorization implies a set of conditional independence (CI) relations. As a simple example, consider the empty DAG on two nodes. A distribution is Markov to this DAG if it satisfies $\bbP(X_1,X_2)=\bbP(X_1)\bbP(X_2)$, which implies that $X_1$ is independent of $X_2$, which we denote by $X_1\independent X_2$. More generally, the Markov property implies for every missing edge a collection of conditional independence relations associated to it which can be read off from the DAG (via \emph{d-separation} criteria~\cite{pearl2014probabilistic}); {see Section~\ref{sec:causal-discovery}.}

\vspace{0.2cm}

\noindent\textbf{Causal discovery:} In many cases, the causal graph $\cG$ is unknown, and only samples of $\bX$ from the joint distribution on the nodes are available. The problem of inferring the underlying causal graph from data is known as \emph{causal discovery}. By the Markov property, missing edges in the graph correspond to CI relations.
If the reverse also holds—an assumption known as \emph{faithfulness} \cite{lauritzen1996graphical,spirtes2000causation}, which we will discuss in detail in Section~\ref{sec:causal-discovery}—then the \emph{adjacencies}, i.e., presence or absence of edges in the DAG, can be inferred from data. In the example in \cref{fig:1a}, we may observe in the data that ice cream sales ($X_2$) is independent of the number of sunburn incidences ($X_3$) conditional on the weather ($X_1$), and therefore infer that there is no edge between $X_2$ and $X_3$ in the underlying causal graph $\cG$. However, note that while some edge directions can be inferred from CI relations under the faithfulness condition (we will for example see in Section~\ref{sec:causal-discovery} that in the 3-node setting $X_1\independent X_2$ and $X_1\notindependent X_2\mid X_3$ imply $1\rightarrow 3 \leftarrow 2$), not all edge directions can be inferred. For example, in the 2-node setting with no CI relations, we cannot distinguish $1\rightarrow 2$ from $1\leftarrow 2$. Thus, with observational data alone, the underlying DAG $\cG$ is only identifiable up to an equivalence class known as the \emph{Markov equivalence class} (MEC)~\cite{verma1990causal}, denoted as $[\cG]$. It is possible to represent $[\cG]$ with a partially directed graph, known as the \emph{essential graph} $\cE(\cG)$ \cite{andersson1997characterization}, which has the same adjacencies as $\cG$ and a directed edge $i\to j$ in $\cE(\cG)$ if and only if it is directed in the same way in all $\cG'\in[\cG]$.
We will discuss algorithms for causal discovery, i.e., for learning the Markov equivalence class of $\cG$, in Section~\ref{sec:causal-discovery}.


\vspace{0.2cm}

\noindent\textbf{Interventional data:} In some settings, we may have access to \emph{interventional} data (also called \emph{perturbational} data in biomedical applications), which can help in directing causal edges. An intervention is defined by a set of target variables $I\subseteq [p]$ and a set of associated \emph{modified mechanisms} $\bbP^I(X_i\mid X_{\Pa_\cG(i)})$ for $i\in I$ resulting in the  \emph{interventional distribution}:
\begin{equation}\label{eq:interv-markov}
    \bbP^I(\bX) = \prod_{i\not\in I} \bbP(X_i\mid X_{\Pa_\cG(i)}) \prod_{i\in I} \bbP^I(X_i\mid X_{\Pa_\cG(i)}). 
\end{equation}
In general, an intervention can result in any modified mechanism $\bbP^I(X_i\mid X_{\Pa_\cG(i)})$. For example, a \emph{do} intervention sets its targeted variable to a specific value, i.e., $\bbP^I(X_i\mid X_{\Pa_\cG(i)})=\delta_{x_i}$, where $\delta_{x_i}$ is the Dirac distribution centered at $x_i$~\cite{meinshausen2016methods}.
Comparing the interventional distribution $\bbP^I$ with the observational distribution $\bbP$ may enable identification of the underlying causal model beyond what is possible from observational data alone.
Returning to the example in \cref{fig:1a}, when we intervene on ice cream sales ($X_2$), e.g., through promotions, we observe that the weather ($X_1$) remains unchanged ($\bbP^I(X_1)=\bbP(X_1)$); from this we can infer that $X_2$ is not upstream of $X_1$. Similarly, by intervening on sunburn incidences ($X_3$), e.g., by applying sun screen, we observe that the weather ($X_1$) remains unchanged and thus that $X_3$ is not upstream of $X_1$, which allows us to fully orient the causal graph.
%
Unlike many other fields, biology benefits from modern techniques that allow interventions to be applied at scale.
Such interventions can take the form of genetic perturbations, e.g., using CRISPR-based methods~\cite{knott2018crispr}, or chemical treatments, e.g., in small-molecule chemical screens~\cite{eggert2006small}.
A genome-wide CRISPR screen can involve thousands of perturbations~\cite{feldman2019optical,replogle2022mapping,huang2025x}, providing the opportunity to address a wide range of causal questions. In Sections~\ref{sec:causal-discovery} and~\ref{sec:crl}, we will discuss how to use interventional data for causal tasks.

\begin{figure}[ht]
    \centering
    \begin{subfigure}{0.3\linewidth}
         \centering
        \includegraphics[width=.667\linewidth]{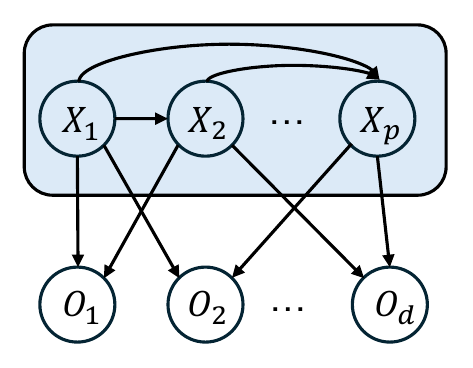}
        \caption{Graphical model representing causal representation learning.}\label{fig:crl}
    \end{subfigure}
    \hfill
    \begin{subfigure}{0.3\linewidth}
        \centering
        \includegraphics[width=.48\linewidth]{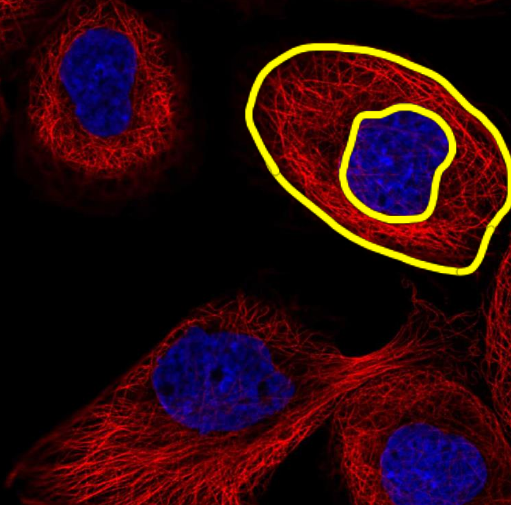}
        \caption{Image of cells (from \href{https://www.proteinatlas.org/ENSG00000126602-TRAP1/subcellular\#human_cell_lines}{Human Protein Atlas}).}\label{fig:2a}
    \end{subfigure}
    \hfill
    \begin{subfigure}{0.3\linewidth}
        \centering
        \includegraphics[width=.5\linewidth]{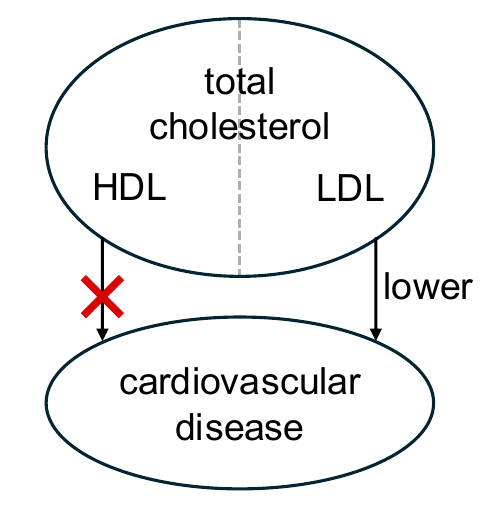}
        \caption{Total cholesterol and cardiovascular disease.}\label{fig:2b}
    \end{subfigure}
    \caption{Illustrative examples of causal representation learning. }
    \label{fig:2}
\end{figure}

\vspace{0.2cm}
\noindent\textbf{Causal representation learning:} In some settings, the causal variables may not be known, and we collect data that do not directly measure the variables of primary interest. For example, we may take microscopy images of cells; each pixel in the image is certainly not a causal variable, but the shape of the cell or the amount and localization of a particular protein in the cell could be a causal variable (\cref{fig:2a}). From indirect measurements we may still be able to identify the underlying causal variables and the rules governing their interactions. 
Consider the setting in \cref{fig:crl}, where the observed variables, denoted as $\bO$, are generated by the underlying causal variables, denoted as $\bX$. The goal of \emph{causal representation learning} (CRL) is to recover $\bX$ as well the causal relations between the variables $\bX$ and the mapping (also known as the \emph{mixing function}) from $\bX$ to $\bO$~\cite{scholkopf2021toward,squires2023linear}. A closely related problem, which can be formulated as a subproblem  
of CRL, is that of \emph{causal feature learning} (CFL), where the goal is to coarsen the observed variables $\bO$, 
to obtain macrovariables $\bX$ 
by partitioning the space of their realizations according to a downstream task \cite{chalupka2017causal}. 
It is worth noting that recovering $\bX$ from $\bO$ is not always achievable, as some applications require finer grained measurements than $\bO$, making it impossible to invert $\bO$ to obtain $\bX$.
For example, relying solely on total cholesterol as measurement in $\bO$ can obscure the causal relation with cardiovascular disease (\cref{fig:2b}). A more fine-grained distinction between high-density lipoprotein (HDL) and low-density lipoprotein (LDL) is necessary, as reducing LDL has been shown to causally lower disease risk, whereas decreasing HDL does not confer the same effect~\cite{spirtes2004causal}. In Section~\ref{sec:crl}, we will discuss theory and methods for CRL.
We note that the problem of CRL considered here differs from the problem of causal discovery in the presence of latent variables, 
where the data directly measure the causal variables (though not all of them) and the main focus is on recovering the causal relationships between the observed causal variables, while in CRL the main focus is on discovering the causal variables from related data. Extensive literature exists on causal discovery with latent variables~\cite{spirtes2000causation,spirtes2001anytime,colombo2014order,ogarrio2016hybrid,jabbari2017discovery,squires2020permutation} which we will not discuss in detail here.

\begin{figure}[ht]
    \centering
\begin{subfigure}{0.3\linewidth}
        \centering
        \includegraphics[width=.9\linewidth]{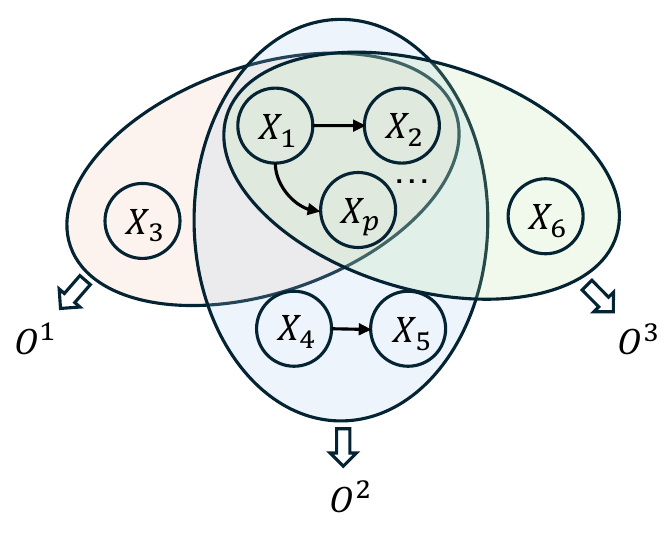}
        \caption{Graphical model representing\\multi-modal causal representation learning.}\label{fig:mdcrl}
    \end{subfigure}
    \begin{subfigure}{0.25\linewidth}
        \centering
        \includegraphics[width=.65\linewidth]{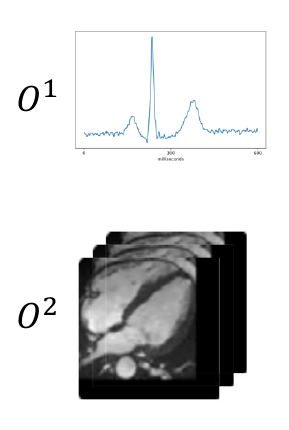}
        \caption{Heart ECGs \\and MRIs.}\label{fig:mdcrl-eg1}
    \end{subfigure}
    \begin{subfigure}{0.3\linewidth}
        \centering
        \includegraphics[width=.48\linewidth]{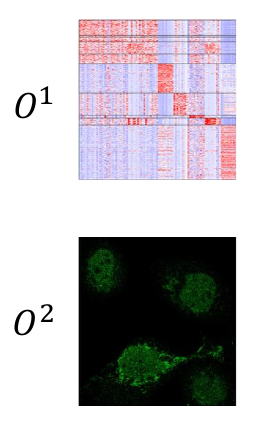}
        \caption{Single-cell gene expression and imaging data. 
        }\label{fig:mdcrl-eg2}
    \end{subfigure}
    \caption{Illustrative examples of multi-modal causal representation learning. }
    \label{fig:3}
\end{figure}

\vspace{0.2cm}
\noindent\textbf{Making use of multi-modal data:} In addition to interventional data, which are a form of multi-modal data, we may also have multiple views or measurements of the system available, with each view providing information on a subset of the causal variables. \Cref{fig:3} shows an example of a graphical model representing this scenario, where three complementary views $\bO^1, \bO^2,\bO^3$ generated by the underlying causal variables $\bX$, are available. The goal in this case is to learn the shared causal variables (e.g., $X_1,X_2,X_p$), modality-specific causal variables (e.g., $X_3,X_4,X_5,X_6$), and the causal relations between them.
For example, clinicians leverage measurements across complementary diagnostic modalities to develop an integrated understanding of the physiological state of a patient. \Cref{fig:mdcrl-eg1} shows an example of two modalities, electrocardiograms (ECGs) containing myoelectric information and cardiac magnetic resonance images (MRIs) containing structural information on the state of the heart of an individual. Similarly, to obtain a more comprehensive understanding of the state of a cell, biologists use sequencing-based assays to measure gene expression and high-resolution imaging to capture the spatial localization of specific proteins (\cref{fig:mdcrl-eg2}). {In Section~\ref{sec:crl}, we will discuss how such multi-modal data can enhance the identification of causal variables and the causal relations among them.}

\vspace{0.2cm}
\noindent\textbf{Optimal design of interventions:} Learning the underlying causal model and data-generating process not only provides a fundamental understanding of the system, but also enables generalization and prediction of a system's behavior under novel conditions. This, in turn, offers a path for manipulating the system toward a desired outcome. In the fibrosis example in \cref{fig:1b}, identifying disease-causal genes/proteins is critical for the development of a therapy.
More generally, although high-throughput perturbational experiments are now feasible in the biomedical sciences \cite{dixit2016perturb,feldman2019optical}, the main challenge lies in the vast space of possible perturbations. It is practically impossible to experimentally test the vast space of drug-like molecules (which is estimated to be of size $10^{24}$~\cite{bohacek1996art}) or to exhaustively perturb the combination of all 20,000 human genes. These huge search spaces create a unique opportunity for computational approaches that can predict the effect of unseen perturbations, allowing virtual screening of perturbations and identification of promising candidates without the need for exhaustive experimental exploration. Such methods have the potential to accelerate therapeutic discovery and inform strategies to modulate cellular states in a targeted manner. In Section~\ref{sec:expdesign}, we will discuss strategies for identifying optimal interventions/perturbations.

\section{Causal discovery.}\label{sec:causal-discovery}
Consider a random vector $\bX$ whose joint distribution $\bbP$ is Markov with respect to a DAG $\cG$ (see Definition~\ref{def:markov}).  Causal discovery is concerned with the problem of inferring $\cG$ given samples of $\bX$. 
The Markov property implies a collection of CI relations that can be fully characterized using \emph{d-separation}~\cite{pearl2014probabilistic}; namely, two nodes $i,j$ are \emph{d-separated} in $\cG$ by a set $S\subseteq [p]\setminus\{i,j\}$, denoted by $i\independent j \mid S$, if all paths\footnote{A path in a DAG $\cG$ is a sequence of nodes such that any two consecutive nodes are adjacent in $\cG$.} connecting $i$ and $j$ are \emph{blocked} by $S$. A path is \emph{blocked} by $S$ if it contains a node $k$ satisfying one of the following conditions:
\begin{enumerate}
    \item $k\notin S$ is a \emph{collider} on the path, i.e., the adjacent nodes $l, h$ on the path satisfy $l\rightarrow k \leftarrow h$, and all its descendants $\Des_\cG(k):=\{\ell\in [p]\mid\exists~\text{a directed path from }k\text{ to }\ell\text{ in }\cG\}$ are not in $S$;
    \item $k\in S$ is not a collider on the path. 
\end{enumerate}
\begin{lemma}
If a distribution $\bbP$ is Markov with respect to a DAG $\cG$, then d-separation implies conditional independence, i.e., $i\independent j\mid S$ in $\cG$ $\Longrightarrow$ $X_i\independent X_j\mid X_S$ in $\bbP$.
\end{lemma}
The proof of this lemma can be found in \cite{verma1990causal,geiger1990identifying}. Essentially, one can apply an inductive argument by first considering three nodes and then extending it to additional nodes via the graphoid axioms \cite{pearl2022graphoids}. 
While d-separation implies conditional independence, the converse does not necessarily hold. The faithfulness condition assumes this reverse implication, thereby allowing one to infer information about $\cG$ from $\bbP$.
\begin{definition}\label{def:faithful}
    A joint distribution $\bbP$ is \emph{faithful} to a DAG $\cG$ if conditional independence implies d-separation, i.e.,  $X_i\independent X_j\mid X_S$ in $\bbP$ $\Longrightarrow$ $i\independent j\mid S$ in $\cG$.
\end{definition}


In other words, the faithfulness assumption guarantees that two nodes that are \emph{d-connected} (i.e., not d-separated) in $\cG$ cannot appear to be conditionally independent in $\bbP$. Thus intuitively, the faithfulness assumption precludes causal effects along different paths to cancel each other out. The Markov condition together with the faithfulness assumption 
allow us to learn about the underlying DAG $\cG$ as long as the correct CI relations $X_i\independent X_j\mid X_S$ were inferred. In the finite-sample regime, the CI relations need to be estimated from the data and thus a stronger form of faithfulness is needed. For example, in the multivariate Gaussian setting, Fisher's z-transform \cite{fisher1924distribution}, i.e., a cutoff on the partial correlations depending on sample size, is used to obtain the CI relations.  

\begin{definition}
    For fixed $\lambda\in [0,1]$, a multivariate Gaussian distribution $\bbP$ is \emph{$\lambda$-strong faithful} to a  DAG $\cG$ if for any nodes $i,j$ in $\cG$ and set $S\subseteq [p]\setminus\{i,j\}$ with $|\corr(X_i,X_j\mid X_S)|\leq\lambda$ it holds that $i\independent j\mid S$ in $\cG$.
\end{definition}
Note that in the infinite-sample setting we can choose $\lambda=0$, which results in the standard faithfulness assumption in Definition~\ref{def:faithful}. 
Note also that the set of distributions violating faithfulness has Lebesgue measure $0$~\cite{spirtes2000causation}, suggesting that faithfulness is a mild assumption.
However, in the finite-sample regime where $\lambda>0$ the set of distributions violating $\lambda$-strong faithfulness no longer has measure $0$.
In particular, \cite{uhler2013geometry}
showed that the measure of $\lambda$-strong-unfaithful distributions can converge to $1$ exponentially in the number of nodes $p$.
This is due to the curvature of the varieties corresponding to faithfulness violations and that thickening these varieties can quickly become ``space-filling''; we illustrate this via the following example on $3$ variables.

\begin{figure}[ht]
    \centering
\begin{subfigure}{0.23\linewidth}
        \centering
        \includegraphics[width=.7\linewidth]{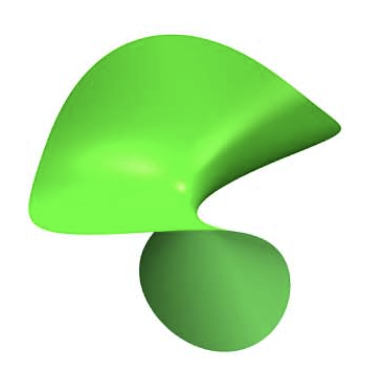}\caption{$\corr(X_1,X_3)=0$.}
    \end{subfigure}
    \hfill
    \begin{subfigure}{0.23\linewidth}
        \centering
        \includegraphics[width=.7\linewidth]{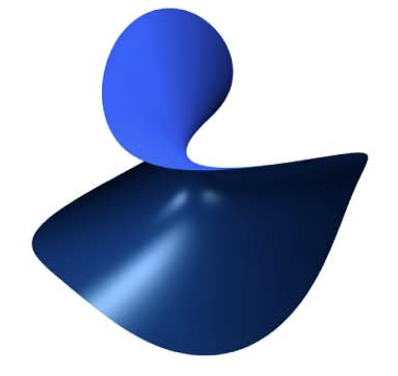}\caption{$\corr(X_1,X_2\mid X_3)=0$.}
    \end{subfigure}
    \hfill
    \begin{subfigure}{0.23\linewidth}
        \centering
        \includegraphics[width=.7\linewidth]{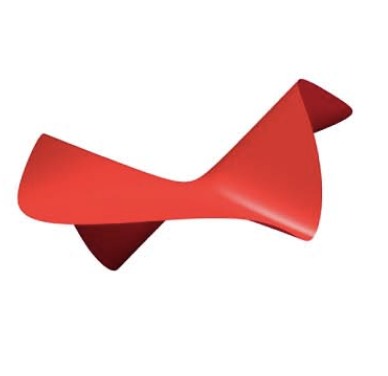}\caption{$\corr(X_2,X_3)=0$.}
    \end{subfigure}
    \hfill
    \begin{subfigure}{0.23\linewidth}
        \centering
        \includegraphics[width=.7\linewidth]{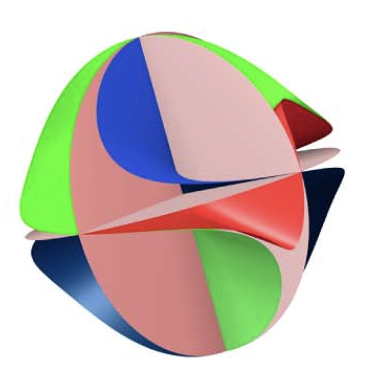}
        \caption{All 6 surfaces}
    \end{subfigure}
    \hfill
    \caption{Surfaces in $\bbR^3$ that correspond to unfaithful distributions for fully connected 3-node linear Gaussian causal models.}
    \label{fig:4}
\end{figure}

\begin{example}
Consider a Gaussian distribution that satisfies the Markov property with respect to the fully-connected DAG $\cG$ with edges $1\to 2, ~1\to 3,~2\to 3$ given by the following linear structural equations:
\begin{align*}
    X_1 & = \epsilon_1,\\
    X_2 & = a_{12} X_1 + \epsilon_2,\\
    X_3 & = a_{13}X_1 + a_{23}X_2 + \epsilon_3,
\end{align*}
where $\epsilon_1,\epsilon_2,\epsilon_3$ are independent standard Gaussians. Since $\cG$ is fully connected, faithfulness requires all 6 partial correlations, $\corr(X_1,X_2)$, $\corr(X_1,X_3), \corr(X_2,X_3), \corr(X_1,X_2\mid X_3), \corr(X_1,X_3\mid X_2), \corr(X_2,X_3\mid X_1)$, to be non-zero. \Cref{fig:4} shows the hypersurfaces of $(a_{12},a_{13},a_{23})^\top$ in $\bbR^3$ that correspond to faithfulness violations. Since $\lambda$-strong faithfulness violations correspond to the by a factor of $\lambda$ ``thickened'' hypersurfaces, it is apparent from the figure that even for small values of $\lambda$, i.e., large sample sizes, the corresponding volume is considerable.
\end{example}

This implies fundamental limitations for causal discovery algorithms that are based on testing many conditional independence relations, since the true distribution needs to be bounded away from all hypersurfaces corresponding to negative CI tests. This motivates the study of algorithms that either (1) rely less on CI testing or (2) perform as few CI tests as possible. In the following, we review various causal discovery algorithms. These are typically grouped into three categories: \emph{constraint-based}, \emph{score-based}, and \emph{hybrid} methods. Constraint-based methods infer the underlying graph $\cG$ by performing CI tests and iteratively pruning DAGs that violate these constraints. In contrast, score-based methods assign a score to each possible DAG, quantifying its fit to the data, and then search for the DAG that maximizes this score. Hybrid methods combine these ideas, for example by restricting the search space using CI relations and then optimizing a score within this reduced space.
For simplicity of the discussion below, we assume that the joint distribution $\bbP$ satisfies the Markov property and faithfulness assumption with respect to $\cG$, meaning that d-separation in $\cG$ is equivalent to CI relation in $\bbP$, and that we have enough samples to fully determine all CI statements in $\bbP$. It is worth noting that many of the algorithms below do not require these assumptions to hold to be correct; see e.g.~\cite{teh2024general} for a characterization of the correctness conditions of constraint-based methods.

\subsection{The PC algorithm.} 
\begin{figure}[ht]
    \centering
    \begin{subfigure}{0.2\linewidth}
        \centering
        \includegraphics[width=.8\linewidth]{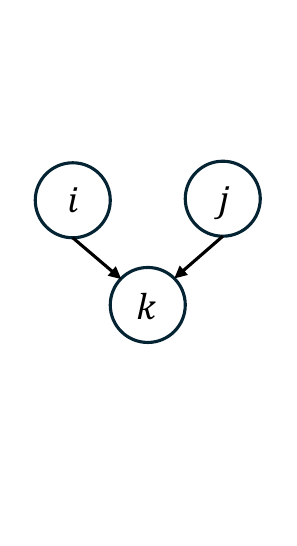}
        \caption{V-structure.}\label{fig:v-structure}
    \end{subfigure}
    \begin{subfigure}{0.7\linewidth}
        \centering
        \includegraphics[width=.8\linewidth]{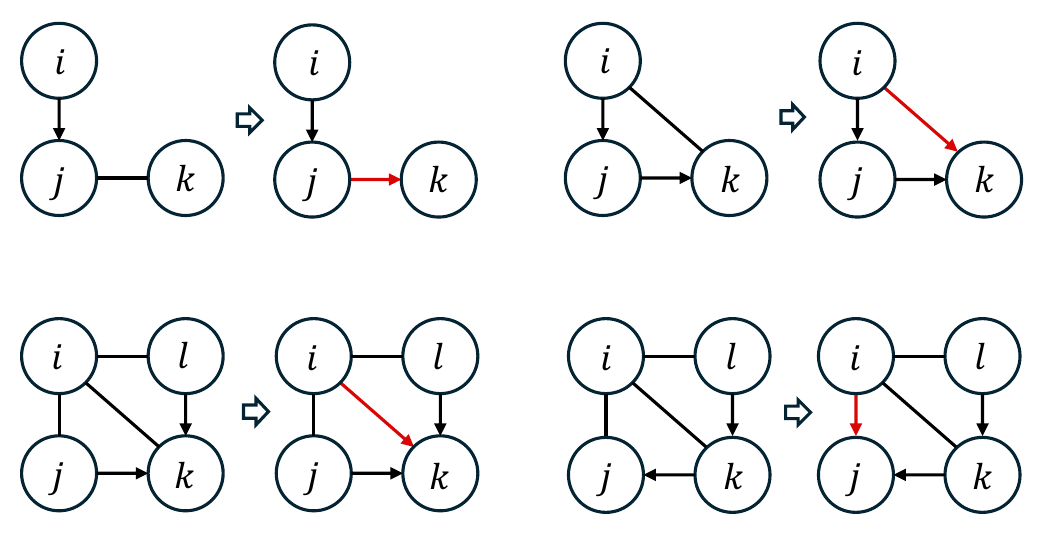}
        \caption{Meek rules (1-4 from left to right and top to bottom).}\label{fig:meek-rule}
    \end{subfigure}
    \caption{V-structure and Meek rules.}
    \label{fig:v-structure-meek-rules}
\end{figure}

This pioneering causal discovery algorithm consists of two steps~\cite{spirtes2000causation}: (1) starting from a fully connected undirected graph, it iteratively removes edges between variables that are conditionally independent given some conditioning set; and (2) it orients edges given the CI relations used in step (1). Step (1) fully identifies the adjacencies in $\cG$.   
To see this, note that two nodes $i,j$ are not adjacent in $\cG$ if and only if they are d-separated by some set $S\subseteq [p]\setminus\{i,j\}$; for example, $S$ can be chosen to be the parents of node $i$, i.e., $\Pa_\cG(i)=\{k\in[p] \mid k\to i\}$, assuming without loss of generality that there is no directed edge from $i$ to $j$. Step (2) of the algorithm orients some edges, first by identifying all \emph{v-structures} in the DAG, i.e., triplets of nodes $(i,j,k)$ where $i,j$ are not adjacent with $i\to k$ and $j\to k$ (\cref{fig:v-structure}), and then applying additional orientation rules known as Meek rules \cite{meek2013causal}. Note that given the node adjacencies inferred in step (1) of the algorithm, the v-structures in $\cG$ are identifiable as triplets of nodes $(i,j,k)$ where $i,j$ are not adjacent and for which there exists a set $S\subseteq[p]\setminus\{i,j\}$ such that $i\independent j\mid S$ and $i\notindependent j\mid S\cup\{k\}$. The Meek rules, shown in \cref{fig:meek-rule}, orient additional edges ensuring the graph remains acyclic and no new v-structures are introduced. Importantly, the Meek rules are complete, meaning that step (2) of the PC algorithm orients all edges that are identifiable, i.e., it outputs the essential graph of $\cG$~\cite{andersson1997characterization}.
%

\subsection{The GAS algorithm.}\label{sec:CCPG}

Let $d$ denote the maximum in-degree of the underlying causal DAG $\cG$. The PC algorithm at most requires $p^{\Omega(d)}$ number of CI tests~\cite{claassen2013learning}, where the main bottleneck lies in the adjacency search, i.e., step (1). A large number of CI tests not only exacerbates the computational burden but also, as reviewed at the beginning of Section~\ref{sec:causal-discovery}, requires strong faithfulness assumptions on the data generating distribution to be bounded away from the hypersurfaces corresponding to negative CI tests. In recent work~\cite{mones2025on}, we provided the following lower bound on the number of CI tests required by any constraint-based causal discovery algorithm.


\begin{theorem}\label{thm:constraint-based-lower-bound}
Given observational data from a distribution that is Markov and faithful to a DAG $\cG$, any algorithm requires at least $\exp(\Omega(s))$ CI tests to verify $\cG\in [\cG]$, where $s$ is the size of the maximal undirected clique\footnote{A clique is an induced subgraph in which all nodes are adjacent.} in the essential graph $\cE(\cG)$.
\end{theorem}
To show this, we proved that for any collection of fewer than $2^s-s-1$ CI tests, one can construct a very similar but different MEC $[\cG']\neq [\cG]$ such that both classes are indistinguishable based
on these CI relations~\cite{zhang2024membership}. Complementing Theorem~\ref{thm:constraint-based-lower-bound}, we also provided a causal discovery algorithm, \emph{greedy ancestral search} (GAS), that matches this lower bound~\cite{mones2025on}. 

\begin{theorem}\label{thm:CCPG-full}
    Given observational data from a distribution that is Markov and faithful to a DAG $\cG$ with $p$ nodes, the GAS algorithm outputs $\cE(\cG)$ using at most $p^{\cO(s)}$ CI tests, where $s$ is the size of the maximum undirected clique in $\cE(\cG)$.
\end{theorem}

\begin{figure}[htbp]
  \centering
  \includegraphics[width=0.49\columnwidth]{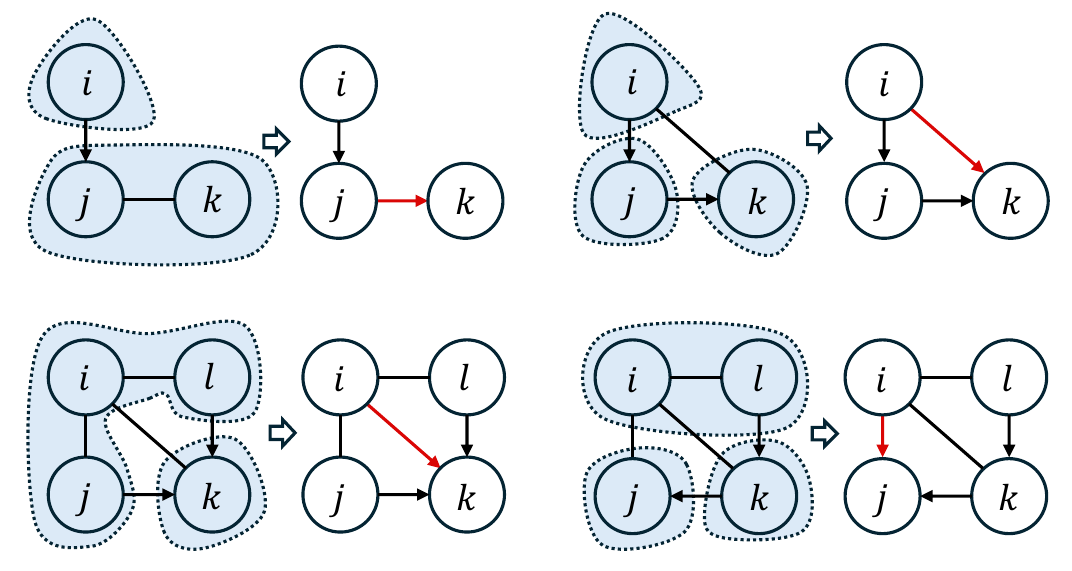}
  \caption{Meek rules (1-4 from left to right and top to bottom) with ancestral partitions indicated in blue.}
  \label{fig:meek-rule-partition}
\end{figure}

Note that since the most downstream node in the maximum undirected clique has an in-degree of at least $d-1$, it holds that $s\leq d-1$; thus the PC algorithm in general performs more CI tests than required. In the following, we will discuss the key ideas of GAS, which build on our previous work~\cite{shiragur2024causal}, where we characterized what can be learned about the Markov equivalence class $[\cG]$ given only a polynomial number of CI tests. To reduce the number of CI tests compared to the PC algorithm, GAS integrates steps (1) and (2); namely, the algorithm focuses on using CI tests to learn ancestral relationships, which can then be used to perform CI tests to uncover adjacencies in a more targeted way. To provide some intuition, note that if we were given all ancestral relationships, i.e., a permutation $\bpi=(\pi_1,\dots,\pi_p)$ of the nodes $[p]$ that is \emph{consistent} with the DAG $\cG=([p], E)$ (i.e., if for $i,j\in[p]$ there is a directed edge $\pi_i\to\pi_j\in E$, then it has to hold that  $i<j$), then a single CI test would be sufficient to determine the presence/absence of an edge:
\begin{equation}\label{eq_min_IMAP}
(\pi_i,\pi_j)\in E \quad \Longleftrightarrow \quad X_{\pi_i} \not\independent X_{\pi_j} \mid X_S, \qquad\textrm{ where } S=\{\pi_1, \pi_2, \dots , \pi_{j-1}\}\setminus\{\pi_i\}.
\end{equation}
Thus, if we were given the correct ordering of the nodes, then a single CI test per edge would suffice. As a consequence, the main difficulty of causal structure discovery is to learn the correct ordering/permutation of the nodes and to identify CI tests that provide ancestral information. Towards this, we analyze the orientation rules, i.e., step (2), in the PC algorithm and note that the ancestral relationships of a DAG are fully identified by its v-structures and Meek rule 1. To see this, consider the four Meek rules in \cref{fig:meek-rule}. In \cref{fig:meek-rule-partition} we indicate the ancestral relationships in the graph before applying each Meek rule by partitioning the nodes into different sets if they are connected by directed edges and grouping nodes that are only connected via undirected edges into the most upstream partition they are connected to. It is apparent in \cref{fig:meek-rule-partition} that only Meek rule 1 provides additional ancestral information beyond transitivity. Based on this insight, we developed two sets of CI tests that allow us to identify the ancestral relationships given by v-structures and Meek rule 1~\cite{mones2025on}. 
To describe these, we use the notion of \emph{descendants} of a node defined above.

\begin{lemma}[CI test for v-structure]
    Consider a DAG $\cG$ with nodes $[p]$. For any three nodes $i,j,k\in [p]$ and set $S\subseteq [p]\setminus \{i,j,k\}$, if $X_i\independent X_j\mid X_S$ and $X_i\notindependent X_j\mid X_{S\cup\{k\}}$, then $i,j\notin \Des_\cG(k)$. 
\end{lemma}

\begin{lemma}[CI test for Meek Rule 1]
    Consider a DAG $\cG$ with nodes $[p]$. Let $S\subseteq[p]$ be a prefix set, i.e., 
     $i\in S$ then all its ancestors (i.e., all nodes with a directed path pointing to $i$) belong to $S$. For any three nodes $i,j,k$ such that $i\in S$ and $j,k\not\in S$, if $X_i\notindependent X_k \mid X_{S\setminus\{i\}}$ and $X_i\independent X_k\mid X_{S\cup\{j\}\setminus\{i\}}$, then $j\notin \Des_\cG(k)$. 
\end{lemma}

GAS iteratively expands the conditioning set $S$ in these two lemmas to learn all ancestral relations. In particular, it maintains a sequence of prefix sets $(S_\ell)_{\ell=1,\dots, L}$, where starting in $S_0=\emptyset$ at each step $\ell$ it obtains $S_\ell$ from $S_{\ell-1}$ by greedily adding elements according to these two lemmas. A detailed description of the algorithm and the full proof of Theorem~\ref{thm:CCPG-full} can be found in~\cite{mones2025on}.


\subsection{The GSP algorithm.}
Beyond constraint-based methods, score-based and hybrid methods assign a score to each possible DAG (or MEC) based on its fit to the data, and might thus rely less heavily on CI tests. In the following, we describe the \emph{greedy sparsest permutation} (GSP) algorithm~\cite{solus2021consistency}, as a representative example of such causal discovery methods. GSP is conceptually related to the GAS algorithm and preceded it. 

A joint distribution $\bbP$ may factorize with respect to multiple DAGs, not just the underlying DAG $\cG$; for example, a distribution where all nodes are independent factorizes with respect to any DAG. Any such DAG is called an \emph{independence map} (I-MAP) of $\bbP$. In fact, in Section~\ref{sec:CCPG} we have already seen how to obtain all \emph{minimal I-MAPs}, i.e., I-MAPs of $\bbP$ where the removal of any edge would result in a new DAG that is no longer an I-MAP of $\bbP$~\cite{verma1990causal}. Namely, for any permutation $\bpi$ of the nodes, define the DAG $\cG_{\bpi}=([p], E_{\bpi})$ by property~(\ref{eq_min_IMAP}). Note that minimal I-MAPs of a DAG $\cG$ may have different number of edges. For a simple example, consider the 3-node DAG $\cG$ with edges $1\to 3$, $2\to 3$. Then the minimal I-MAPs with respect to the permutations $1<2<3$ and $2<1<3$ are equal to $\cG$ and thus have two edges, while the minimial I-MAPs with respect to all other permutations are fully connected. We proved that under the Markov and faithfulness assumptions the sparsest I-MAP must be Markov equivalent to $\cG$~ \cite{raskutti2018learning}:



\begin{theorem}
Given a joint distribution $\bbP$ that is Markov and faithful with respect to a DAG $\cG$, any I-MAP $\cG'$ of $\bbP$ such that $\cG'\not\in [\cG]$ (if it exists) must contain strictly more adjacencies than $\cG$.
\end{theorem}


This result directly suggests the \emph{sparsest permutation} (SP) algorithm~\cite{raskutti2018learning}: enumerate all permutations $\bpi$, obtain the corresponding minimal I-MAP $\cG_{\bpi}$ using CI tests in $\bbP$, then use the number of edges as a score function to return the sparsest DAG.
However, this procedure is clearly computationally prohibitive, as the number of possible permutations is $p!$. 

The greedy sparsest permutation (GSP) algorithm mitigates this by greedily searching over the space of permutations~\cite{solus2021consistency}: At each step $i$, GSP maintains a permutation $\bpi^i$ and its corresponding minimal I-MAP $\cG_{\bpi^i}$. It then searches over all DAGs in the Markov equivalence class $[\cG_{\bpi^i}]$ for some DAG $\cG'$ that is not a minimal I-MAP of $\bbP$. This search can be executed by repeatedly flipping \emph{covered}\footnote{An edge $i\rightarrow j$ in a DAG $\cG$ is \emph{covered} if the parents of $j$ are exactly the parents of $i$ plus $i$ itself, i.e., $\Pa_\cG(j)=\Pa_{\cG}(i)\cup\{i\}$.} edges, as guaranteed by the construction of a Chickering sequence~\cite{chickering2002optimal} based on Meek’s conjecture~\cite{meek2013causal}. Since $\cG'$ is not a minimal I-MAP of $\bbP$ (it is an I-MAP since $\cG'\in[\cG_{\bpi^i}]$), there must exist edges that can be removed to obtain $\cG''$ which is a minimal I-MAP of $\bbP$. GSP then takes the permutation consistent with $\cG''$ as the new permutation $\bpi^{i+1}$, with $\cG''$ as its corresponding minimal I-MAP. We showed that GSP is guaranteed to return the correct MEC; a more detailed description of the algorithm and the proof of Theorem~\ref{thm:GSP} can be found in~\cite{solus2021consistency}.
\begin{theorem}\label{thm:GSP}
    Given observational data from a distribution that is Markov and faithful to a DAG $\cG$, the GSP algorithm outputs the essential graph $\cE(\cG)$.
\end{theorem}

Building on GSP, \cite{lam2022greedy} introduced a family of causal discovery algorithms called GRaSP, which employ a novel traversal strategy in the space of DAGs, referred to as ``tuck''. They showed that the lowest tier of GRaSP is equivalent to GSP, while higher tiers require weaker faithfulness assumptions for correctness. In fact, \cite{teh2024general} showed that the correctness condition required by SP is among the weakest of all causal discovery algorithms. However, since SP requires enumerating all permutations, it incurs substantial computational cost. This suggests an important trade-off between correctness condition and computational efficiency that remains to be better understood.

\subsection{Interventional data.}
Without additional assumptions~\cite{peters2017elements}, from observational data alone it is only possible to identify the MEC of the underlying causal graph. As motivated in Section~\ref{sec:intro}, interventional data may improve the identifiability of the underlying causal DAG. However, similar to the observational setting, a faithfulness assumption is required to ensure that the effects of interventions do not de-generate and can be used for causal discovery. \cite{tian2013causal} introduced the following interventional faithfulness assumption based on the marginal distribution of the targeted variables.
\begin{definition}\label{int_faithful}
   Given a joint distribution $\bbP$ that is Markov and faithful with respect to a DAG $\cG$. An intervention $I$ defined by the modified mechanisms $\bbP^I(X_i\mid X_{\Pa_\cG(i)})$ for $i\in I$ satisfies the \emph{interventional faithfulness condition} if the interventional distribution $\bbP^I(\bX)$ defined in Equation (\ref{eq:interv-markov}) satisfies $\bbP^I(X_j)\neq \bbP(X_j)$ for all nodes $j\in \cup_{i\in I}\Des_\cG(i)$.
\end{definition}
It follows directly from the definition of the interventional distribution in Equation (\ref{eq:interv-markov}) that $\bbP^I(X_j)=\bbP(X_j)$ for all nodes $j\not\in \cup_{i\in I}\Des_\cG(i)$. 
Therefore, interventional faithfulness guarantees that we can identify nodes that are downstream of an intervention. Algorithms building upon this intuition can identify additional edge orientations in $[\cG]$~\cite{hauser2012characterization}. 
In particular, we used this idea to extend GSP to the interventional setting~\cite{wang2017permutation,yang2018characterizing} and we also considered the problem of learning from interventional data without performing any CI tests~\cite{mazaheri2025faithfulness}.

\subsection{Application to learning gene regulatory networks.}
A concrete application of these algorithms arises in the inference of gene regulatory networks. Here, the variables $\bX$ correspond to the expression levels of individual genes, and the causal graph $\cG$ specifies the regulatory relationships between them. For example, a transcription factor $X_i$ regulates another gene $X_j$ (denoted $i \rightarrow j$ in $\cG$) by binding to the cis-regulatory element of that gene on the DNA. With current single-cell RNA-seq technologies, each data point corresponds to the gene expression measurement of a single cell across all genes~\cite{baysoy2023technological}. 
Such single-cell measurements can be coupled with CRISPR-based techniques~\cite{knott2018crispr} to perturb individual genes through knock-out, repression, or activation. This technology, known as Perturb-seq~\cite{dixit2016perturb}, allows simultaneous measurement of the expression of all genes as well as the perturbation that was performed on the cell, providing large-scale perturbational datasets.

In \cite{solus2021consistency}, we analyzed the first Perturb-seq dataset on bone-marrow–derived dendritic cells~\cite{dixit2016perturb}; in particular, we applied PC and GSP to the $992$ observational samples, and we used the $13,534$ interventional samples across $8$ gene deletions for evaluating the output of these algorithms. In the analysis, we restricted the dimensionality to $p=24$, focusing on transcription factors known to regulate a variety of genes, including one another~\cite{garber2012high}. More recently, in \cite{mones2025on} we applied GAS, which is significantly faster due to the small number of CI tests performed, to single-cell gene expression data generated by the SERGIO simulator \cite{dibaeinia2020sergio}. This allowed us to compare the inferred structures directly against the ground-truth DAG. The experiments involved $2{,}700$ cells with expression profiles on $p=100$ genes. 
In~\cite{wang2017permutation,yang2018characterizing}, we extended GSP to the interventional setting, and applied it to \cite{dixit2016perturb}.
Beyond gene expression data, in \cite{wang2017permutation,yang2018characterizing}, we also applied the interventional versions of GSP to study a protein signaling network based on a mass spectrometry dataset consisting of $5{,}846$ measurements of phosphoprotein and phospholipid levels in primary human immune cells~\cite{sachs2005causal}; interventions in this setting correspond to chemical reagents that inhibit or activate specific signaling proteins. Furthermore, in~\cite{belyaeva2021dci}, we proposed a method to directly learn differences in gene regulatory mechanisms across conditions, and we applied this approach to two single-cell gene expression datasets containing perturbational data for validation~\cite{datlinger2017pooled,dixit2016perturb}.

\section{Going beyond observed causal variables through causal representation learning.}\label{sec:crl}
In Section~\ref{sec:causal-discovery}, we considered the setting where the causal variables $\bX=(X_1,\dots,X_p)^\top$ of interest are directly observed. In many cases, however, we only have data on variables $\bO=(O_1,\dots,O_d)^\top$ that do not directly measure $\bX$, as discussed in Section~\ref{sec:intro}. In this section, we describe approaches to learn $\bX$ from $\bO$. In particular, we will consider three different settings. In Section~\ref{sec:obs-crl}, we consider the single-modality setting, where we have access to samples from $\bO=\bbf(\bX)$, with latent causal variables $\bX$ drawn from a distribution $\bbP$ that is Markov with respect to a causal DAG $\cG$, and $\bbf$ an unknown mixing function. 
In Section~\ref{sec:int-crl}, we consider the interventional setting, where we have access to samples from $\bO^I=\bbf(\bX^I)$ for $K$ different interventions $I\in\{I_1,\dots,I_K\}$ (note that $I=\varnothing$ reduces to the previous setting), where $\bX^I$ is drawn from the interventional distribution $\bbP^I$, which is obtained from a distribution $\bbP$ that is Markov with respect to a causal DAG $\cG$, and $\bbf$ is an unknown mixing function that remains fixed across interventions. 
Finally, in Section~\ref{sec:mm-crl}, we consider the setting with $M$ partially overlapping modalities, $\bO^1=\bbf^1(\bX), \dots, \bO^M=\bbf^M(\bX)$, where the causal variables $\bX$ are drawn from a distribution $\bbP$ that is Markov with respect to a causal DAG $\cG$, and $\bbf^1, \dots, \bbf^M$ are unknown modality-specific mixing functions. Note that this includes the case where $\bO^i$ is a function of just a subset of the latent variables $\bX$.

In each subsection below, we will consider the problem of \emph{identifiability}, namely whether it is possible to recover the underlying causal variables $\bX$ and their relationships $\cG$, as well as algorithms to do so. For simplicity, we will assume throughout access to sufficient samples from $\bO$, $\bO^I$, $I\in\{I_1, \dots, I_K\}$ and $\bO^1,\dots,\bO^M$ to fully determine their distributions. Note that in general, we cannot achieve full identifiability. 
For example, there is a trivial non-identifiability corresponding to renaming variables: if we simultaneously permute the entries of $\bX$ and the function $\bbf$ according to the same permutation $\bpi=(\pi_1,\dots,\pi_p)$, we obtain the same observed variables: $\bbf(\bX)=\bbf_{\bpi}(\bX_{\bpi})$, $\bX_{\bpi}=(X_{\pi_1},\dots,X_{\pi_p})^\top$ are the permuted variables and the permuted mixing function $\bbf_{\bpi}$ corresponds to permuting the input–output mapping of $\bbf$. Similarly, one can apply element-wise affine transformations to both $\bX$ and $\bbf$ without changing $\bO$.
Thus, at most we can identify the underlying causal model up to an \emph{equivalence class}. 

\subsection{Causal representation learning from single-modality data.}\label{sec:obs-crl}

The identifiability problem is difficult as it encompasses both disentanglement (to identify $\bX$) and causal discovery (to identify $\cG$). 
Traditionally, in the disentanglement literature, the latent factors are assumed to be independent, and it is known that identifying them is not possible without additional assumptions on the data-generating process~\cite{hyvarinen1999nonlinear}. We are interested in the more general setting where the latent causal variables $\bX$ may be related and we aim to discover not only  $\bX$ but also their relationships $\cG$. Since the traditional disentanglement problem is a special case, $\bX$ is unidentifiable without additional assumptions. 
In the following, we describe an approach that utilizes asymmetries in $\bbP$ to learn $\bX$~\cite{welch2024identifiability}. Towards this, we consider the following three assumptions:
\begin{assumption}\label{ass:3-1}
    The mixing function $\bbf$ is linear and invertible, i.e., there is a full-column rank matrix $\bF\in\bbR^{d\times p}$ such that $\bO=\bF\bX$.
\end{assumption}

\begin{assumption}\label{ass:3-2}
     The factors of the joint distribution $\bbP$ of $\bX$ are specified by
     \[
     X_i = h_i(X_{\Pa_\cG(i)}) + \epsilon_i,\quad \forall i\in [p],
     \]
     where each $h_i$ is a twice continuously differentiable, non-linear function that captures the dependence of $X_i$ on its parents, and each $\epsilon_i\sim \cN(0,\sigma_i^2)$ corresponds to an exogenous noise variable that is mutually independent and mean-zero Gaussian.
\end{assumption}

\begin{assumption}\label{ass:3-3}
    For any $i\in [p]$ and any non-zero vector $\beta\in\bbR^{|\Pa_\cG(i)|}$, the random variable $\partial^2_{\beta,\beta} h_i(X_{\Pa_\cG(i)})$ is not always zero.
\end{assumption}

Assumption~\ref{ass:3-1} constrains the space of possible mixing functions $\bbf$.
We note that we do not assume prior knowledge of the dimension $p$ of the latent causal variables. Since $\bbf$ is linear and invertible, $p$ can be identified from the intrinsic dimension of the joint distribution of $\bO$.  Note also that we can assume without loss of generality that the linear mapping is zero-centered, since as discussed at the beginning of Section~\ref{sec:crl} that we will at best be able to identify $\bX$ up to element-wise affine transformations. 
The nonlinear
causal model with additive Gaussian noise in Assumption~\ref{ass:3-2} has been a popular choice in the causal discovery literature due to its flexibility, identifiability properties (in the fully observed setting), and benign statistical sample complexity requirements~\cite{peters2012identifiability,scholkopf2012causal,rolland2022score,zhu2023sample}. Assumption~\ref{ass:3-3} functions similarly to the faithfulness assumption, ensuring that the causal effect of a parent on a child is non-degenerated. The class of nonlinear
causal models with additive Gaussian noise implies an asymmetric relationship between causes and effects, which can be utilized to infer causal relations and fully identify the underlying causal model in the setting where all causal variables are observed~\cite{rolland2022score}. The following lemma summarizes the key property. Its proof can be found in \cite{rolland2022score}.

\begin{lemma}\label{lemma}
    Let $\bJ(\bx)$ denote the Jacobian matrix of $\bX$ at $\bx$ with $ij$-th entry given by $\bJ(\bx)_{ij}=\nabla_{x_i}\nabla_{x_j} \log \bbP(\bx)$. The $i$-th diagonal element of the Jacobian matrix has zero variance, i.e., $\var(\bJ(\bX)_{ii})=0$, if and only if node $i$ is a \emph{leaf node} in $\cG$, i.e., $\Des_\cG(i)=\varnothing$.
\end{lemma}


Although we are unable to obtain $\bJ(\bX)$ since $\bX$ is not directly measured, a similar result holds despite the unknown mixing function $\bF$: Let $\bJ_\bO(\bo)$ denote the Jacobian matrix of $\bO$ at $\bo$, defined analogously to $\bJ(\bx)$.  Let $\bF^\dagger = (\bF^\top\bF)^{-1}\bF^\top$ denote the Moore-Penrose inverse. Since $\bJ(\bF^\dagger\bo)=\bF^\top \bJ_\bO(\bo)\bF$, then by Lemma~\ref{lemma} it is possible to infer leaf-node information from $\bF^\top \bJ_\bO(\bo)\bF$. 
We showed in \cite{welch2024identifiability} that if we solve for $\bF$ by maximizing the number of zero diagonal entries in $\var(\bF^\top\bJ_\bO( \bO)\bF)$, we obtain exactly the number of leaf nodes in $\cG$. More precisely, we showed the following result.

\begin{lemma}\label{lm:obs-crl}
    Let the matrix $\hat{\bF}$ be obtained by solving
    \begin{equation}\label{eq:lm-obs-crl}
    \begin{aligned}
    \min_{{\bF}\in\bbR^{d\times p}} &\quad\left\| \var\Big(\diag\big(\bF^\top\bJ_\bO( \bO)\bF\big)\Big)\right\|_0, \\
    \text{such that} &\quad\text{rank}({\bF})=p.
    \end{aligned}
    \end{equation}
Denoting by $\bJ_{\hat{\bX}}$ the Jacobian matrix of $\hat{\bX}$, then it follows that $\hat{\bX}=\hat{\bF}^\dagger\bO$  satisfies
    \begin{equation*}
    \hat{X}_i = 
    \begin{cases}
    \text{linear}(X_{non-leaf})\quad&\text{if }\var(\bJ_{\hat{\bX}}(\hat{\bX})_{ii})\neq 0,\\
    \text{linear}(X)\quad \quad&\text{if }\var(\bJ_{\hat{\bX}}(\hat{\bX})_{ii})= 0,
    \end{cases}
    \end{equation*}
    where the number of nodes $i\in [p]$ such that $\var(\bJ_{\hat{\bX}}(\hat{\bX})_{ii})=0$ equals to the number of leaf nodes in $\cG$. 
\end{lemma}
We note that all Jacobians in Lemma~\ref{lm:obs-crl} can be computed based on the observed samples from $\bO$. In addition, we do not need to know $p$ apriori, since the constrained optimization problem in \cref{eq:lm-obs-crl} can be interpreted as solving for a full-column rank matrix such that the $l_0$ norm is maximized.  We showed in \cite{welch2024identifiability} that this optimization problem can be equivalently formulated as a quadratically constrained quadratic program and efficiently solved by off-the-shelf numerical solvers~\cite{marchand2002cutting}. In summary, Lemma~\ref{lm:obs-crl} provides an approach for iteratively identifying leaf nodes as a linear combination of all variables in its own and upstream layers. This leads to the following identifiability ``up to upstream layers'', where $layer(k)$ is defined as the set of all nodes whose longest path to a leaf node is of length $k$. 


\begin{theorem}\label{thm:peeling}
    Under Assumptions~\ref{ass:3-1},~\ref{ass:3-2}, and~\ref{ass:3-3}, given sufficient samples of $\bO$ the latent causal variables $\bX$ are identifiable up to their upstream layers, i.e., we can learn $\hat{\bX}$ from $\bO$ such that:
    \begin{equation*}
        \hat{\bX} =  \bP_{\bpi}\bC\bX,
    \end{equation*}
    where $\bP_{\bpi}\in\bbR^{p\times p}$ is a permutation matrix, and $\bC\in\bbR^{p\times p}$ is a constant matrix with non-zero diagonal entries and $\bC_{ij}=0$ for all $i,j$ such that $i\in layer(k)$ and $j\in\cup_{l\leq k}layer(l)$. 
\end{theorem}

The full proof of Theorem~\ref{thm:peeling} and a detailed description of the algorithm can be found in~\cite{welch2024identifiability}.

\subsection{Causal representation learning from interventional data.}\label{sec:int-crl}

We next consider the setting where we have access to interventional data. Like for causal discovery, this will result in stronger identifiability results. We here consider the setting where each intervention $\{I_1,\dots, I_K\}$ has a unique target $i$ among the causal variables $\bX$, but the target variable is unknown since $\bX$ is unknown. In addition, we assume that for every causal variable, there is at least one intervention that targets it. Moreover, we make the following two assumptions. 
\begin{assumption}\label{ass:3-8}
    The interior of the support of $\bbP$ is a non-empty subset of $\bbR^p$. The mixing function $\bbf$ is a full-column rank polynomial, i.e., there exists some integer $s$, a full-column rank matrix $\bF\in\bbR^{d\times (p+\dots+p^s)}$ such that $\bO=\bF(\bar{\otimes}\bX, \bar{\otimes}\bX^2,\dots,\bar{\otimes}\bX^s)^\top$, where $\bar{\otimes}\bX^{r}$ denotes the size-$p^r$ row-vector with degree-$r$ polynomials of $\bX$ as its entries.
\end{assumption}

\begin{assumption}\label{ass:3-9}
    \emph{Linear interventional faithfulness} holds for all interventions $I\in\{I_1, \dots, I_K\}$; i.e., let $i$ denote the target of $I$ and let $\Ch_\cG(i)=\{j\in[p]\mid i\in\Pa_\cG(j)\}$ denote the children of $i$ in $\cG$. Then for every $j\in\{i\}\cup\Ch_\cG(i)$ such that $\Pa_\cG(j)\cap \Des_\cG(i)=\varnothing$, it holds that $\bbP(X_j+C^\top X_S) \neq \bbP^I(X_j+C^\top X_S)$ for any constant vector $C\in\bbR^{|S|}$, where $S=[p]\setminus(\{j\}\cup\Des_{\cG}(i))$.
\end{assumption}

Assumption~\ref{ass:3-8} allows us to extend linear mixing in Assumption~\ref{ass:3-1} to polynomial mixing: the support with non-empty interior guarantees that
we can identify the dimension $p$ of $\bX$, and the full-rank polynomial assumption
ensures that we can search for $\bbf$ (and consequently $\bX$) in a constrained subspace~\cite{ahuja2023interventional}. 
We do not need to impose any parametric constraints as in Assumption~\ref{ass:3-2}, since interventions allow us to exploit the principle of invariance rather than asymmetries in the observational distribution to identify the causal relations \cite{buhlmann2020invariance}.
However, to apply the principle of invariance, we must assume that interventions induce changes in the system.
Assumption~\ref{ass:3-9} extends the interventional faithfulness assumption in Definition~\ref{int_faithful} from the causal discovery setting with observed causal variables to the setting where we only observe a linear mixing of the causal variables. In this setting, a stronger condition is needed to ensure that the effect of intervening on a causal variable $X_i$ not only affects its children, but that the effect will not be canceled out through linear combinations with other causal variables that are not downstream of $X_i$.
Note that this condition only needs to hold for $i$ itself and the most upstream child of $i$, which may be much smaller than the set of all children of $i$.

Under these assumptions, we can show that we can identify the causal variables and their relationships by detecting marginal changes made by interventions. To provide some intuition, consider the easier setting where $K=p$, i.e., we have exactly one intervention per latent causal variable.  
Based on the intuition provided in the previous paragraph, we can reduce the polynomial mixing to a linear mixing from $\bbR^p$ to $\bbR^p$. Thus we consider the case where we have access to $\bO \in \bbR^p$, which is a linear transformation of $\bX$. 
Note that for a source node $i$ of $\cG$, $\bbP(X_i) \neq \bbP^I(X_i)$ if and only if the target of $I$ is $i$.
By enforcing that the learned $\hat{\bX}$ is of the form $\hat{X}_i = C^\top\bO$ and that it satisfies $\bbP(\hat{X}_i)\neq \bbP^I(\hat{X}_i)$ for exactly one $I\in \{I_1,\dots,I_p\}$, then 
Assumption~\ref{ass:3-9} guarantees that $\hat{X}_i$ can only be an affine transformation of a source node and that the particular intervention $I$ corresponds to intervening on this source node. The argument is as follows: (1) If $C_j \neq 0$ for a non-source node $j$, then let $j$ be the most downstream node with $C_j \neq 0$, in which case $\bbP(\hat{X}_i)\neq \bbP^I(\hat{X}_i)$ for at least two interventions targeting $j$ and its most downstream parents in $\Pa_\cG(j)$, which is a contradiction;  (2) If $C_{i_1}\neq 0$ and $C_{i_2}\neq 0$ for two source nodes $i_1,i_2$, then $\bbP(\hat{X}_i)\neq \bbP^I(\hat{X}_i)$ for two interventions targeting $i_1$ and $i_2$, which is a contradiction.
In general, we can apply this argument to identify all interventions in $I_1,...,I_K$ that target source nodes of $\cG$. Then using an iterative argument, we can identify all interventions that target source nodes of the subgraph of $\cG$ after removing its source nodes. This procedure results in the ancestral relations between the targets of $I_1,...,I_K$. 
This argument holds more generally even when causal variables are targeted by multiple interventions. Denoting by $\Anc_\cG(j):=\{i\in[p]\mid j\in\Des_\cG(i)\}$ the set of \emph{ancestors} of a node $j$ in $\cG$ and by $\cT\cS(\cG)$ its \emph{transitive closure}, i.e., $i\to j \in\cT\cS(\cG)$ if and only if $i\in\Anc_\cG(j)$, we showed the following result~\cite{zhang2023identifiability}.

\begin{theorem}\label{thm:3-10}
    Under Assumptions~\ref{ass:3-8} and~\ref{ass:3-9}, given sufficient samples from $\bO,\bO^{I_1},\dots,\bO^{I_K}$ the transitive closure  $\cT\cS(\cG)$ and the targets of the interventions $I_1, \dots , I_K$ are identifiable up to a  permutation $\bpi$ of the variables $[p]$. If in addition for every edge $i\to j\in \cG$, for any constants $c,(d)_{k\in S}\in\bbR$ there is $$X_i\not\independent X_j+cX_i\mid (X_{\Pa_\cG(j)\setminus (S\cup\{i\})}, \{X_k + d_kX_i\}_{k\in S}),$$ where $S=\Pa_\cG(j)\cap\Des_\cG(i)$, then the full causal graph $\cG$ is identifiable up to a permutation $\bpi$ of the variables $[p]$.
\end{theorem}

In general, we cannot identify beyond the transitive closure of $\cG$, since the effect of a direct edge may be explained by the transitive effects of multiple edges. 
DAGs with the same transitive closure can span a spectrum of sparsities; for example, a complete graph and a line graph with the same topological ordering have the same transitive closure. The additional assumption in the theorem can be seen as an additional interventional faithfulness condition and guarantees identifiability of $\cG$ (up to a permutation of the nodes). While in this case we can associate each latent causal variable with the interventions that target it (i.e., we can interpret the causal latent variables) and we can fully identify the causal structure among the latent causal variables, identification up to a permutation means that we cannot identify $\bX$ in an element-wise fashion. 




\vspace{0.2cm}
\textbf{Application to learning gene regulatory networks.} We next discuss the implications of our identifiability results in the context of large-scale Perturb-seq screens~\cite{norman2019exploring}. Given infinite high-dimensional single-cell transcriptomic readouts from a whole-genome Perturb-seq screen $\bO,\bO^{I_1},\dots,\bO^{I_K}$, Theorem~\ref{thm:3-10} guarantees that we can identify the interventions that act on the same latent node, the ancestral relationships among the intervention targets, and—under the additional assumption in the theorem—the exact causal structure.
This means that we can identify the number of latent causal variables (which we can interpret as the gene programs of a cell), which genes belong to the same program, as well as the full regulatory relationships between the programs. 


In~\cite{zhang2023identifiability}, we turned these theoretical results into a practical autoencoding variational Bayes framework to estimate the latent causal representation from interventional data using maximum mean discrepancy. 
By applying our computational framework to a Perturb-seq study~\cite{norman2019exploring}, we tested its ability to identify gene programs and regulatory networks between programs, as well as on the task of predicting the effect of unseen combinatorial interventions. The Perturb-seq dataset~\cite{norman2019exploring} contains 8,907 unperturbed cells (i.e., samples from $\bO$) and 99,590 perturbed cells that underwent CRISPR activation \cite{gilbert2014genome} targeting one or two out of 105 genes (samples from $\bO^{I_1}$,\dots,$\bO^{I_K}$ with $K = 217$). In~\cite{liu2025learning}, we extended this framework to be able to incorporate prior knowledge on the gene regulatory network and applied it to a subset of a Perturb-seq experiment on K562 cells with $279$ perturbations and more than $200$ cells per perturbation~\cite{replogle2022mapping}. 

\subsection{Causal representation learning from partially overlapping multi-modal data.}\label{sec:mm-crl}
While the interventional setting considered in Section~\ref{sec:int-crl} can be seen as multi-modal, with each intervention being a modality that provides a distinct view on the full causal system, we now consider the general multi-modal setting where each modality in $\bO^1,\dots,\bO^M$, with $\bO^m\in\bbR^{d_m}$ for $m\in [M]$, is not necessarily interventional and may provide information only on a subset of the causal variables. In this case, we have $M$ different mixing functions $\bbf^1, \dots , \bbf^M$, one for each modality. Let $\bX_{\cL}$ with $\cL\subseteq [p]$ denote the set of latent causal variables that are shared across the $M$ modalities, and we denote by $\bX_{\cL^m}$, $m\in[M]$ the modality-specific latent causal variables, i.e., $\cL^1\cup\dots\cup\cL^M=[p]\setminus \cL$. We assume that the modality-specific latent causal variables are disjoint, i.e., $\cL^{i}\cap\cL^{j}=\varnothing$ if $i\neq j$. 
Moreover, we make the following three assumptions.



\begin{assumption}\label{ass:3-14}
    The mixing functions $\bbf^1, \dots , \bbf^M$ are linear and invertible, i.e., for each modality $m\in[M]$ there is a full-column rank matrix  $\bF^m\in\bbR^{d_m\times(|\cL|+|\cL^m|)}$ such that $\bO^m = \bF^m(\bX_{\cL}^\top, \bX_{\cL^m}^\top)^\top$. 
\end{assumption}

\begin{assumption}\label{ass:3-16}
The factors of the joint distribution $\bbP$ of $\bX$ are specified by
    \[
    X_i = A_i^\top X_{\Pa_\cG(i)}  + \epsilon_i, 
    \]
    where $A_i\in\bbR^{|\Pa_\cG(i)|}$ and the exogenous noise variables $\epsilon_i$, $i\in[p]$ are mutually independent, zero-mean, unit variance, non-degenrate, non-symmetric and pairwise different to each other and to the flipped versions, i.e., they satisfy $\epsilon_i\overset{d}={\epsilon_j}$ or ${\epsilon_i}\overset{d}{=} -{\epsilon_j}$ if and only if $i=j$.
\end{assumption}

\begin{assumption}\label{ass:3-15}
    The underlying causal DAG $\cG$ satisfies the following conditions: there is no edge between $\cL$ and $\cL^m$ for any $m\in [M]$ and there is no edge between $\cL^{i}$ and $\cL^{j}$ for any $i\neq j$.
\end{assumption}

Similarly to the assumptions in Section~\ref{sec:obs-crl} and~\ref{sec:int-crl}, Assumption~\ref{ass:3-14} ensures that we can identify the dimension $p$ of $\bX$ and search for the mixing functions (and consequently $\bX$) in a constrained subspace. 
Assumption~\ref{ass:3-16} 
allows us to extend the identifiability results of linear ICA \cite{comon1994independent,eriksson2004identifiability} to our multi-modal setup to identify the joint distribution.
In particular, the assumption of pairwise different error distributions allows for “matching” the distributions across modalities to identify the ones corresponding to the shared latent space. Non-symmetry accounts for the sign-indeterminacy of linear ICA when matching the distributions. Assumption~\ref{ass:3-15} implies that the shared causal variables do not depend on modality-specific ones, and that modality-specific causal variables from one modality do not depend on modality-specific causal variables from other modalities, although causal relations between modality-specific variables of the same modality are allowed. 
Under these assumptions, we showed the following result, namely that one can recover the distribution of the exogenous noise variables and the mapping from these variables to the observed variables up to a permutation, as well as identify the  set of exogenous noise variables corresponding to the shared causal variables $\bX_\cL$.

\begin{theorem}
   Let $\bI_p$ denote the identity matrix of dimension $p$. Let $\bO\in\mathbb{R}^{d_1+\dots+d_M}$ denote the random vector obtained by stacking $\bO^1, \dots , \bO^M$, and similarly, let $\bF\in\bbR^{(d_1+\dots+d_M)\times p}$ denote the matrix obtained by stacking $\bF^1,\dots,\bF^M$ such that $\bO=\bF\bX$, where the variables in $\bX$ are ordered as $(\cL, \cL^1, \dots , \cL^M)$.
   Under Assumptions~\ref{ass:3-14}, \ref{ass:3-16}, and \ref{ass:3-15}, given sufficient samples from $\bO^1, \dots , \bO^M$ we can identify the number of shared latent causal variables $|\cL|$ and we can write $\bO=\hat{\bB}\hat{\pmb{\epsilon}}$, where we can identify the  matrix $\hat{\bB}$ and joint distribution $\hat{\pmb{\epsilon}}\sim\hat{\bbP}$ as follows:
    \[
    \hat{\bB} = \bF(\bI_p-\bA)^{-1}\bP_{\bpi},\qquad \hat{\bbP} = \bbP_{\bpi},
    \]
    where 
    $\bP_{\bpi}\in\bbR^{p\times p}$ is a permutation matrix 
    and $\bbP_{\bpi}$ is the joint distribution of $\epsilon_{\pi_1},\dots,\epsilon_{\pi_P}$. 
\end{theorem}

With the following additional assumptions we can obtain identifiability results on the structure of the shared causal graph (among the latent causal variables that are shared across modalities).

\begin{assumption}\label{ass:3-18}
    For each shared latent node $\ell\in \cL$, there exist two distinct observed variables $\bO_i^*, \bO_j^*$ (can belong to any of the $M$ modalities) that depend only on $X_\ell$.
\end{assumption}
\begin{assumption}\label{ass:3-19}
    For any two subsets $D\subseteq [d_1+\dots+d_M]$ and $L\subseteq \cL$ and any matrices $\tilde{\bA}\in\bbR^{p\times p}$ and $\tilde{\bF}\in\bbR^{(d_1+\dots+d_M)\times p}$ that have the same sparsity pattern as $\bA$ and $\bF$, it holds that $\textrm{rank}((\bF(\bI_p-\bA)^{-1})_{D,L})\geq \textrm{rank}((\tilde{\bF}(\bI_p-\tilde{\bA})^{-1})_{D,L})$.
\end{assumption}

Assumption~\ref{ass:3-18} is a sparsity condition on the concatenated mixing matrix $\bF$; Assumption~\ref{ass:3-19} guarantees that no configuration of edge parameters coincidentally yields low rank. Under these additional assumptions, we showed the following identifiability result in~\cite{sturma2023unpaired}.

\begin{theorem}
    Under Assumptions~\ref{ass:3-14},~\ref{ass:3-16},~\ref{ass:3-15},~\ref{ass:3-18}, and~\ref{ass:3-19}, from sufficient samples $\bO^1, \dots , \bO^M$ the shared causal variables $\bX_\cL$ and the shared causal graph $\cG_\cL$ (defined by the submatrix $\bA_{\cL}$) are identifiable up to permutation and scaling, i.e., we can identify
        \[
    \hat{\bX}_\cL = \bP_{\bpi}\bC\bX_\cL,\quad \hat{\bA}_\cL = \bP_{\bpi}\bC\bA_\cL\bC^{-1}\bP_{\bpi}^{-1},
    \]
    where $\bP_{\bpi}\in\bbR^{|\cL|\times |\cL|}$ is a permutation matrix, and $\bC\in\bbR^{|\cL|\times |\cL|}$ is an invertible diagonal matrix.
\end{theorem}



\vspace{0.2cm}
\textbf{Application to multi-modal integration, translation, and disentanglement of biomedical data.}
We next discuss the implications of our identifiability results in the context of biomedical data together with practical algorithms for learning $\bX$ from $\bO$. We first consider the setting where all latent causal variables $\bX_\mathcal{L}$ are shared across modalities, i.e., $\mathcal{L}=[p]$. We developed various practical approaches based on autoencoders, where modality-specific encoders and decoders are used to map between observed data from each modality $\bO^m, m\in [M]$, and the shared latent space $\bX_{\mathcal{L}}$~\cite{yang2019multi,yang2021multi,radhakrishnan2023cross,zhang2024partially}. When paired data is avaialble across modalities, i.e., we have access to data from the joint distribution $(\bO^1, \dots , \bO^M)$, a constrastive loss can be used in the latent space to align the different modalities. In~\cite{radhakrishnan2023cross}, we applied this approach to construct a holistic representation of cardiovascular state based on two modalities, $\bO^1$ being heart ECG data and $\bO^2$ being heart MRI data. However, in many biomedical applications obtaining a measurement is destructive to the system and thus paired measurements are not available. For example, it is only possible to measure a cell via sequencing or imaging modality, but not both. In \cite{yang2019multi}, when different (unpaired) modalities are generated from a shared latent representation, we proved that the problem of computing a probabilistic coupling $\bX_{\mathcal{L}}$ between marginals of different modalities $\bO^1,\dots , \bO^M$ is equivalent to learning multiple uncoupled autoencoders that embed to a given shared latent distribution.  In \cite{yang2021multi}, we applied this framework to integrate single-cell RNA-seq and chromatin imaging data, which cannot be measured in the same cell, to identify distinct subpopulations of human naïve CD4+ T cells poised for activation~\cite{zheng2017massively}.The advantage of using autoencoder architectures is their generative nature, which allows analyzing a related task, namely whether and how well one modality could be translated to another, in particular if one modality is cheaper and/or easier to obtain. Our work also suggested for the first time that cheap chromatin imaging may contain sufficient information to translate to more expensive and laborious RNA-seq measurements at single-cell resolution. 

In order to obtain the most complete picture of a causal system, it is critical to be able to integrate data of different modalities and understand what information is not shared between modalities. Our identifiability results for partially overlapping multi-modal data suggested that we could go beyond a fully shared latent space and identify information that is modality-specific. 
In \cite{zhang2024partially}, we proposed a computational framework that automatically learns partial information sharing across modalities via an autoencoder with a partially overlapping latent space (for two modalities this latent space corresponds to $(\bX_\mathcal{L}, \bX_{\mathcal{L}^1}, \bX_{\mathcal{L}^2})$). We applied this method to paired scRNA-seq and scATAC-seq data (SHARE-seq)~\cite{ma2020chromatin}, paired scRNA-seq and surface protein data (CITE-seq)~\cite{gayoso2021joint}, and large-scale multiplexed single-cell imaging datasets, such as the Human Protein Atlas~\cite{regev2017human}. In addition to multi-modal data integration and translation, this work provides the first computational framework in the biomedical domain for disentangling information that is shared between different data modalities from information that can only be obtained from a specific modality, a task that is critical for experimental design and the selection of modalities.

\section{Causal experimental design.}\label{sec:expdesign}

We next consider the problem of experimental design in causal systems, where data is collected in an active fashion over multiple rounds either from the observational distribution, from different interventions, or other modalities. This is a relatively nascent area, and we consider two settings: 
the problem of optimal design of interventions, where in each round we can decide which interventions to perform, 
and the more general problem of optimal design of modalities, 
where in each round we can decide from which modality to collect data. 
By adaptively designing experiments taking the current dataset into account, it should be possible to achieve a desired goal more efficiently, with less amount of data, as compared to a passive design.

\begin{figure}[!h]
    \centering
    \includegraphics[width=0.5\linewidth]{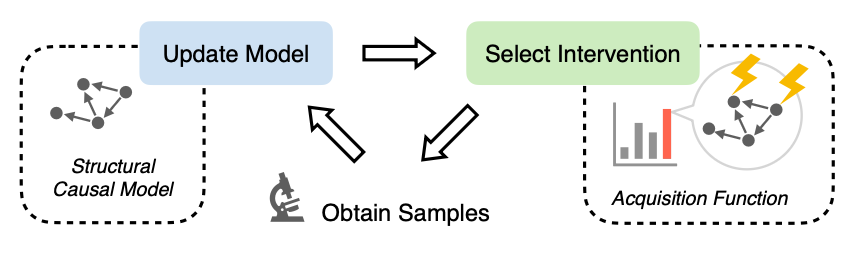}
    \caption{Illustration of iterative intervention design.}
    \label{fig:des-int}
\end{figure}

\vspace{0.2cm}
\textbf{Optimal design of interventions.} 
\Cref{fig:des-int} illustrates the process of experimental design of interventions: The learned perturbation prediction model is iteratively updated over $T$ rounds with the newly collected data. In round 1, a random subset of interventions is selected for which experiments are performed. These experiments, together with data from the observational distribution, are used as warm-up to obtain a predictive model of intervention effects $g_1: I\rightarrow \bbP^I$. 
In round $t+1$, let $\cI_t$ denote the interventions performed so far and let  $g_t$ denote the current model. The interventions $\cI_{t+1}\setminus \cI_t$ for the next batch of experiments are chosen so as to maximize an \emph{acquisition function} $a_t:I\rightarrow \bbR$, which ranks all possible interventions. The problem of optimal design of interventions is relevant for the setting where all causal variables are observed considered in Section~\ref{sec:causal-discovery} as well as the causal representation learning setting considered in Section~\ref{sec:crl}. 

The predictive model $g_t$ is chosen depending on the application of interest. For example, in \cite{agrawal2019abcd} we adopted a Bayesian approach, in which a structural causal model $\cG$ is learned to model the effect of an intervention, with $\cG$ sampled from the Bayesian posterior over all possible DAGs given the current data; in  \cite{zhang2023active}, we  used a linear additive Gaussian causal model on the observed causal models together with shift interventions; and in~\cite{he2025morph} we used discrepancy-based variational autoencoders instead. Similarly, the acquisition function $a_t$ is tailored to the goal of the experiments. For example, in \cite{agrawal2019abcd}, we considered the problem of learning a function of the underlying causal graph (e.g., the set of descendants of a target node) subject to design constraints such as limits on the number of samples and rounds of experimentation, and we used mutual information between this function and the collected samples as the acquisition function. In \cite{zhang2023active}, we considered the problem of achieving a desired target mean for the interventional distribution, and we used output-weighted variance as the acquisition function. We applied this framework to a semi-synthetic experiment to learn genetic interventions that achieve a target mean. In \cite{zhang2021matching}, we considered the same goal but assumed that one can gather infinite data per intervention and used a structure-based acquisition function. We extended this work in \cite{shiragur2023meek} to consider several goals including learning the orientation of edges of a specific set. 
Finally, in~\cite{he2025morph}, we considered the goal of learning a generalizable function $g:I\rightarrow\bbP^I$ for all possible interventions $I$ and experimented with multiple types of uncertainty-based acquistion functions. We applied this framework to the genome-wide Perturb-seq data on K562 cells~\cite{replogle2022mapping} and estimated the number of interventions needed to learn a generalizable intervention effect model.

\vspace{0.2cm}
\textbf{Optimal design of modalities.} 
Technological developments in the past decades have led to an explosion of data of different modalities in the biomedical sciences. Different modalities come with different cost and information. For example, heart ECG data is much more prevalent and cost-effective than heart MRI data, but in unlike MRI data, ECG data is believed to contain only limited structural information. In future work, it will be critical to build on the methods for learning and disentangling shared and modality-specific information discussed in Section~\ref{sec:mm-crl} to develop principled approaches to decide which modality to prioritize given a particular downstream task. 

\vspace{0.2cm}
There is an extensive literature on experimental design, spanning areas such as Bayesian optimization, bandits, reinforcement learning, and uncertainty quantification. 
However, it is not well understood how to incorporate causal information so that experimental design guides data collection in a way that both benefits and contributes to accelerating the discovery of the underlying causal mechanisms. This is a nascent area with many open problems that are of great relevance for the biomedical sciences, where experiments are often performed iteratively and an important end goal is to obtain causal/mechanistic understanding.

\section*{Acknowledgments.}
This review forms the basis for an Invited Section Lecture at the International Congress of Mathematicians 2026. 
We thank all former and current members of the Uhler Lab as well as our collaborators for the many fruitful discussions that have collectively shaped our outlook and made this work possible.

\bibliographystyle{plain}
\bibliography{example_references}
\end{document}